\documentclass[11pt]{article}

\usepackage{graphicx}
\usepackage{amsmath}
\usepackage{amssymb}
\usepackage[skip=2pt]{caption}
\usepackage{subcaption}
\usepackage{float}
\usepackage[hyphens]{url}
\usepackage{hyperref}
\usepackage{geometry}
\title{CAD-to-CT Registration of Cylindrical Objects\\via Ellipse-Based Axis Estimation}

\author{
Aleksander Ogonowski\thanks{Email: aleksander.ogonowski@ncbj.gov.pl},
Miko{\l}aj Mrozowski,
Daniel Wi{\c{e}}cek,
Arkadiusz {\'C}wiek,\\
Konrad Klimaszewski,
Rafa{\l} Mo{\.z}d{\.z}onek,
Adam Padee,
Lech Raczy{\'n}ski,\\
Piotr Wasiuk,
Wojciech Wi{\'s}licki\\
\small Department of Complex Systems, National Centre for Nuclear Research\\
\small ul. So{\l}tana 7, 05-400 Otwock-{\'S}wierk, Poland
\and
Micha{\l} Matusiak,
S{\l}awomir Wronka\\
\small ImagineRT sp.\ z o.o., National Centre for Nuclear Research\\
\small ul. So{\l}tana 7, 05-400 Otwock-{\'S}wierk, Poland
}

\date{}

\begin{document}
\maketitle

\begin{abstract}
Accurate registration of CAD models to CT scans is essential for establishing ground truth geometry in volumetric imaging. Obtaining reliable object masks is of growing importance in machine learning settings; as recent architectures grow more capable, huge datasets are required to fully utilise their capabilities. Traditional intensity-based methods fail when CT grayscale values lack calibration references, while point-based algorithms (e.g., ICP, RANSAC) require feature correspondence unavailable between idealized CAD geometry and noisy volumetric CT data.

We propose a two-stage geometric registration method for cylindrical objects (ionization chambers) that takes advantage of the distinctive geometric features of the objects. First, we estimate the 3D rotation axis by detecting elliptical cross-sections across CT slices, fitting ellipses to edge-detected contours, and performing PCA on the fitted ellipse centers after RANSAC outlier removal. Second, we voxelize the CAD model, orient it along the detected axis, and maximize volumetric overlap with the CT scan through translational adjustment.

This approach achieves robust registration with tilt and orientation errors below $0.1^\circ$ without intensity calibration or feature matching. Once registered, the aligned CAD model provides ground truth geometry for applications including machine learning-based object localization and automated analysis in industrial CT workflows.
\end{abstract}

\section*{Introduction}

Accurate volumetric characterization of ionization chambers from Computed Tomography (CT) scans is essential for quality control in radiation dosimetry. It mainly involves determining the chamber's internal cavity volume and wall thickness, which directly influence its dosimetric response --- even small geometric deviations can introduce systematic measurement errors~\cite{almond1999aapm}.

\begin{figure}[ht]
\centering
\includegraphics[width=0.7\textwidth]{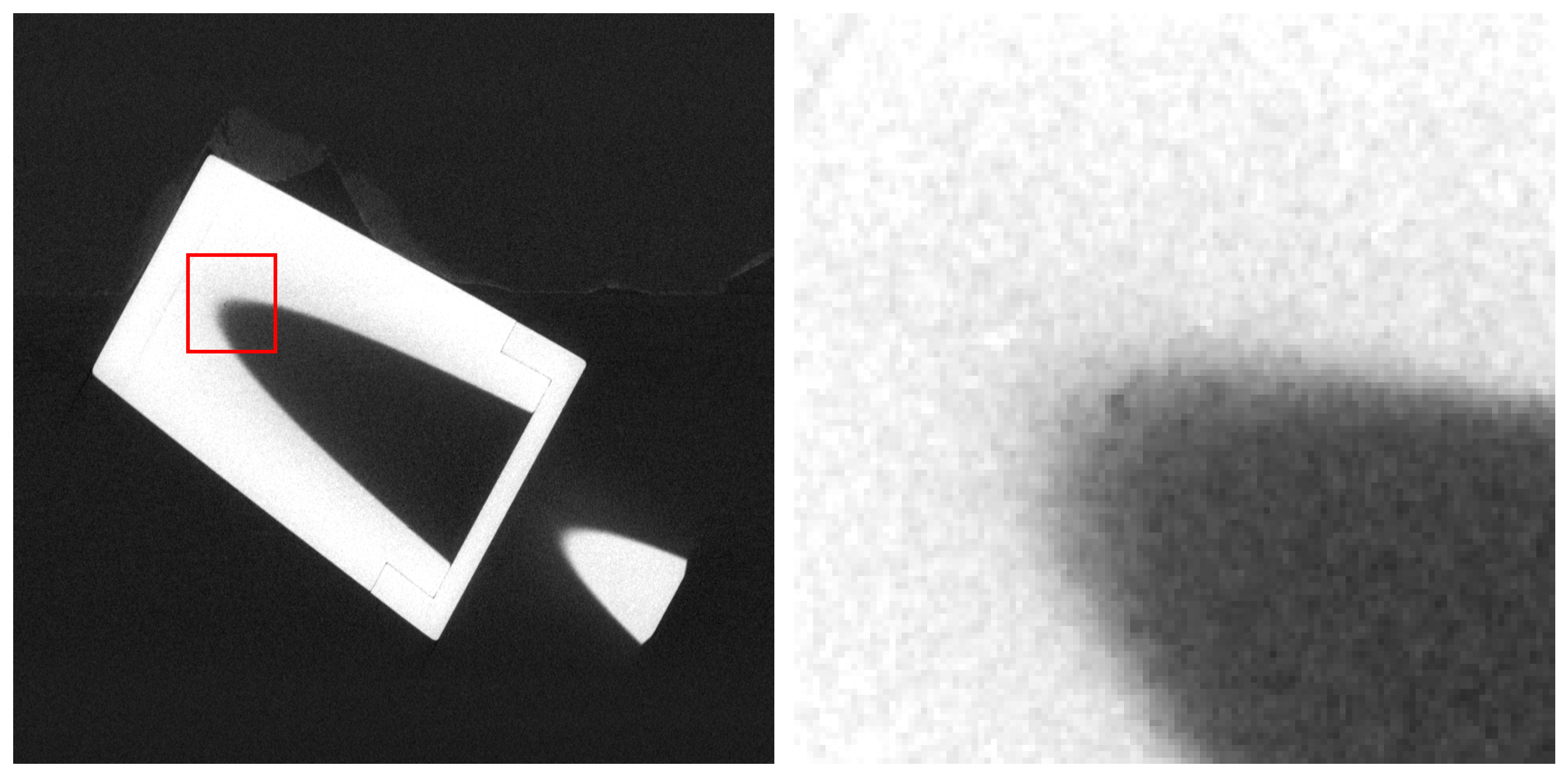}
\caption{Challenge of threshold-based segmentation in CT data: (left) full CT slice showing ionization chamber with region of interest marked in red, (right) magnified view of chamber boundary. The gradual intensity transition and lack of sharp edges make manual threshold selection ambiguous, motivating the need for geometry-based registration to establish ground truth.}
\label{fig:threshold_problem}
\end{figure}

The central challenge in automated segmentation of ionization chambers from CT data is the absence of ground truth masks. Industrial CT grayscale values represent relative X-ray attenuation and vary with scanner calibration, beam energy, and reconstruction parameters~\cite{kruth2011computed}, while artifacts such as beam hardening and metal-induced streaking~\cite{barrett2004artifacts} further complicate boundary localization. As illustrated in Figure~\ref{fig:threshold_problem}, chamber boundaries exhibit gradual intensity transitions rather than sharp edges, making threshold-based segmentation inherently ambiguous --- different threshold choices can shift boundary localization by several voxels~\cite{muller2013computed}, which for small ionization chambers exceeds the sub-voxel accuracy required for reliable dosimetric characterization. Supervised machine learning could address this, but requires labeled training data~\cite{litjens2017survey,ronneberger2015unet} --- precisely what our method is designed to provide.

We address this by leveraging physical reference measurements: selected chambers have undergone high-precision dimensional metrology, and CAD models constructed from these measurements faithfully represent the manufactured geometry. Registering these CAD models to the corresponding CT scans yields voxelized ground truth masks that enable volumetric measurements, tolerance validation, and labeled training data for machine learning-based segmentation. However, this registration is non-trivial --- intensity-based methods~\cite{maes1997multimodality} fail on uncalibrated CT volumes, and point-based techniques such as ICP~\cite{besl1992method} require surface extraction through thresholding, reintroducing the same boundary localization errors~\cite{muller2013computed}.

We propose a geometry-based registration approach that circumvents these limitations by exploiting the cylindrical symmetry of the object. Ionization chambers are manufactured on lathes, which by construction produces rotationally symmetric geometry --- making cylindrical symmetry a reliable prior for this class of instruments. The key observation is that axial CT slices through a tilted cylinder yield elliptical cross-sections whose centers shift linearly across slices --- the slope of this trajectory directly encodes the tilt and azimuth of the rotation axis. Our method consists of two stages. First, we estimate the 3D rotation axis from these elliptical cross-sections: ellipses are fitted to edge-detected contours, their centers across multiple slices form a 3D point cloud, and PCA on the RANSAC inlier set yields the axis orientation without intensity calibration or manual landmarks. Second, we voxelize the CAD model~\cite{nooruddin2003simplification}, orient it along the recovered axis, and maximize volumetric overlap with the CT scan through translational optimization~\cite{zitova2003image}.

We validate our approach through two complementary experiments. First, we evaluate the method on simulated CT data generated directly from the CAD model, where the exact object position and orientation are known by construction --- this allows quantitative assessment of both rotational and translational registration accuracy against the ground truth used to produce the simulation. Second, we apply the method to real ionization chamber CT scans, where per-stage ground truth is unavailable; registration accuracy is assessed indirectly by comparing predicted cavity volumes and wall geometry with measurements from a coordinate measuring machine (CMM)~\cite{ferrucci2015towards}, used here as the geometric ground truth reference.

The main contributions of this work are:
\begin{itemize}
\item A two-stage registration framework robust to intensity threshold selection: (1) sub-voxel 3D axis estimation by fitting ellipses to CT cross-sections and recovering tilt and azimuth from the linear trajectory of their centers across slices --- with results independent of the exact threshold value, and (2) translational alignment via volumetric overlap maximization.
\item A practical dual-validation pipeline: quantitative accuracy assessment on simulated CT with exact ground truth, and end-to-end volumetric validation on real chambers against CMM measurements --- directly enabling automated CT data labelling for supervised segmentation.
\end{itemize}

\section*{Related Work}

\subsection*{Registration of CAD Models to Measurement Data}

Registration of CAD models to measurement data is a common task in manufacturing metrology and quality control. Early approaches focused on aligning CAD models to CMM data using point-based methods~\cite{besl1992method}. The Iterative Closest Point (ICP) algorithm and its variants have been widely applied to register CAD surfaces to measured point clouds~\cite{rusinkiewicz2001efficient}. However, these methods require extracting point sets from both modalities and assume correspondence between surface features, which is problematic for CT data due to the scanner- and material-dependent nature of intensity boundaries.

\subsection*{Industrial CT for Dimensional Metrology}

The use of CT for dimensional measurement of manufactured parts has grown significantly~\cite{kruth2011computed,dewulf2013sense}. A key challenge is the lack of standardized intensity calibration, which makes boundary determination dependent on operator-selected thresholds~\cite{carmignato2012accuracy}. Various threshold selection methods have been proposed, including the ISO 50\% method~\cite{dewulf2013sense}, gradient-based approaches~\cite{tan2016comparison}, and material-specific calibrations~\cite{carmignato2012accuracy}, but all require prior knowledge of material properties or dedicated reference measurements --- knowledge that is difficult to obtain reliably when scanning multi-material assemblies or when scanner parameters vary between acquisitions.

M{\"u}ller et al.~\cite{muller2013computed} demonstrated that threshold choice significantly affects dimensional measurements, with errors up to several voxels for complex geometries. Our approach circumvents this problem by using geometric constraints rather than intensity thresholding for initial alignment.

\subsection*{Geometric Feature Detection for Registration}

\subsubsection*{Cylinder Fitting}

Fitting cylinders to 3D data through circular or elliptical cross-sections has been explored in coordinate metrology and reverse engineering. Luk{\'a}cs et al.~\cite{lukacs1998faithful} presented methods for fitting quadric surfaces including cylinders by analyzing planar cross-sections from laser range data, assuming known scan plane orientations. Shakarji~\cite{shakarji1998least} developed least-squares cylinder fitting algorithms operating on discrete 3D probe measurements from CMMs, where individual contact points with known spatial coordinates are fitted directly. Forbes~\cite{forbes2006least} provided comprehensive treatments of geometric element fitting for dimensional metrology using tactile probe and optical scanning data.

However, these methods rely on surface points directly acquired with known spatial coordinates. CT data provides volumetric grayscale intensities without explicit surface definition --- extracting surface points requires intensity thresholding, which reintroduces scanner-dependent boundary errors~\cite{tan2016comparison}.

\subsubsection*{Other Geometric Primitives}

For 3D point cloud registration, feature-based methods encode local geometric relationships. Point pair features~\cite{drost2010model} and 4-point congruent sets~\cite{aiger2008line} enable robust matching without requiring initial alignment. Fast Point Feature Histograms (FPFH)~\cite{rusu2009fpfh} describe local surface geometry through normal distributions. Global optimization variants of ICP~\cite{yang2015go} have improved registration robustness by avoiding local minima. Cylinder and plane primitives have also been fitted to laser-scanned point clouds for facility reconstruction~\cite{rabbani2006automatic} and pipe detection~\cite{nurunnabi2012robust}.

All these methods assume direct surface measurements where points correspond unambiguously to physical boundaries. Applying them to CT requires intensity-based surface extraction, which introduces the same threshold-dependent errors discussed above --- making feature descriptors relying on surface normals~\cite{rusu2009fpfh} particularly unreliable in uncalibrated CT data.

\subsection*{Intensity-Based Registration Methods}

In medical imaging, multi-modal registration typically aligns CT with MRI, PET, or other imaging modalities~\cite{maes1997multimodality,pluim2003mutual}. These methods rely on intensity-based similarity metrics such as mutual information, which assume consistent statistical relationships between modalities. Such assumptions do not hold when registering binary CAD geometry to uncalibrated CT grayscale volumes. Industrial CT inspection methods~\cite{buratti2018applications} similarly rely on either known object positioning or initial alignment --- neither of which is available in our setting.

\subsection*{Volumetric Overlap Maximization}

Overlap-based registration metrics have been used in medical image registration~\cite{klein2010elastix}, typically for mono-modal alignment where intensity distributions are comparable. Zitov{\'a} and Flusser~\cite{zitova2003image} reviewed various registration similarity measures including overlap-based metrics, but focused on 2D image registration. Our approach adapts this idea to the cross-modal CAD-to-CT case by first establishing orientation through geometric constraints, making the overlap score robust to absolute intensity values.

\subsection*{Summary and Positioning}

Existing registration methods for metrology either rely on surface point extraction with scanner-dependent thresholds or assume intensity-based correspondence. Our method is the first to exploit cylindrical symmetry through ellipse trajectory analysis for axis estimation whose results are independent of the exact threshold value, followed by volumetric overlap refinement. This approach is particularly suited to industrial CT metrology where intensity calibration is unavailable and geometric priors are strong.

\section*{Method}

\subsection*{Overview}

The registration method consists of two stages: (1) \textbf{rotation axis estimation} through ellipse trajectory analysis with RANSAC-based outlier removal and PCA-based axis extraction, and (2) \textbf{translational alignment} through volumetric overlap maximization. The method takes as input a CT volume $V_{\text{CT}} \in \mathbb{R}^{N_x \times N_y \times N_z}$ with isotropic voxel size $s_v$ [mm], and aligns to it a reference CAD model represented as a triangular mesh $M_{\text{CAD}}$. The registration currently targets only the chamber interior fill and wall geometry; electrode stems and connector elements are excluded from the CAD mesh used for alignment, though the framework can straightforwardly be extended to include additional components.

\begin{figure}[ht]
\centering
\includegraphics[width=0.7\textwidth]{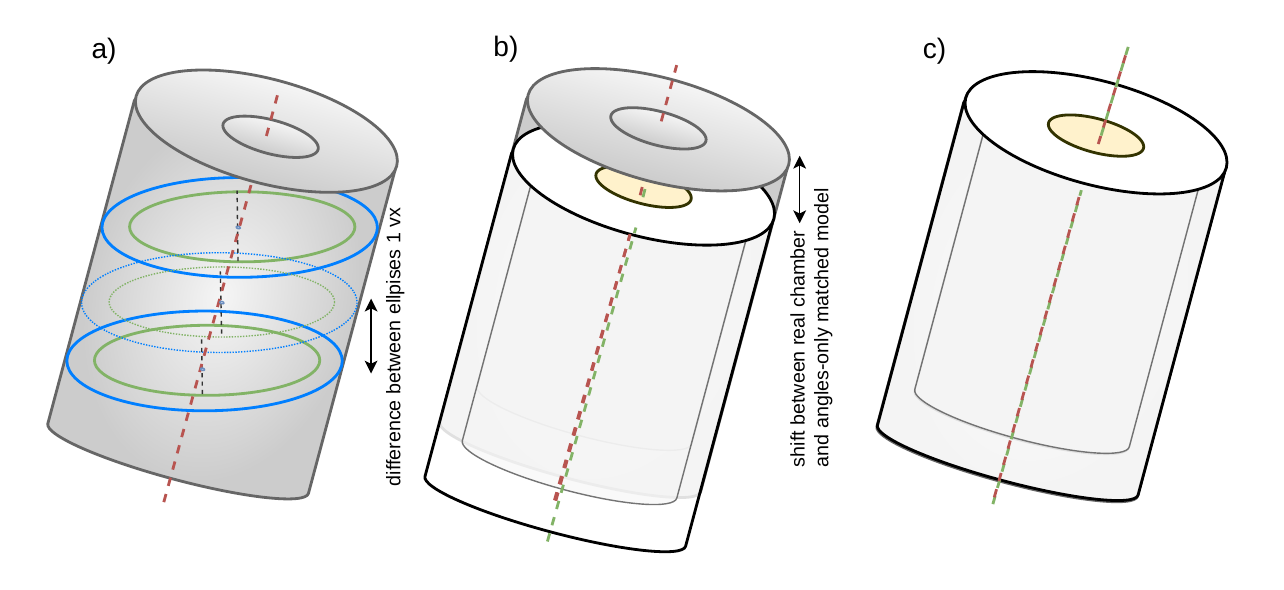}
\caption{Overview of the CAD-to-CT registration pipeline: (a) rotation axis estimation from ellipse centers fitted to inner (green) and outer (blue) chamber contours across axial slices; (b) translational alignment along the rotation axis; (c) final fine alignment result with the CAD model accurately positioned along the recovered axis.}
\label{fig:pipeline_overview}
\end{figure}

\subsection*{Data Preprocessing and Slice Extraction}

\subsubsection*{CT Volume Representation}

\begin{figure}[ht]
\centering
\includegraphics[width=0.7\textwidth]{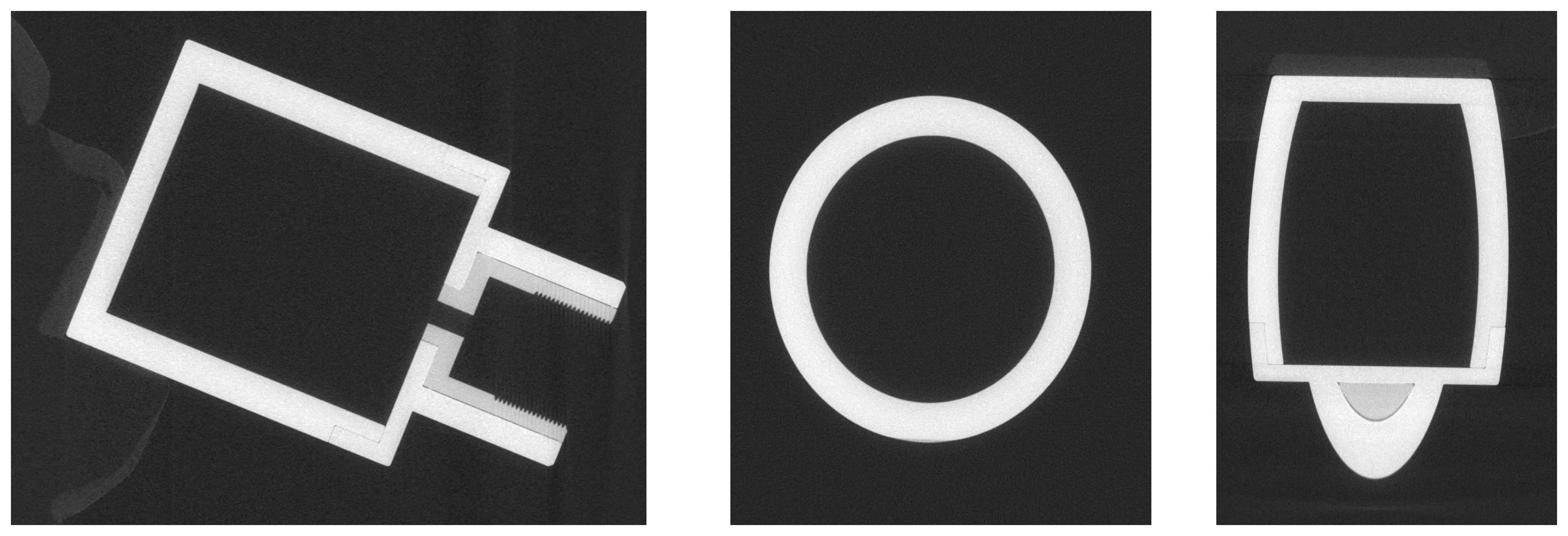}
\caption{Three orthogonal views of the ionization chamber CT scan showing axial, coronal, and sagittal slices.}
\label{fig:orthogonal_slices}
\end{figure}

The CT scan is represented as a discretized 3D scalar field $V_{\text{CT}}(i,j,k) : \mathbb{Z}^3 \to \mathbb{R}$, where $(i,j,k)$ are voxel indices and $V_{\text{CT}}(i,j,k)$ represents grayscale intensity (linear attenuation coefficient). Physical coordinates are obtained by scaling with the isotropic voxel size $s_v$, i.e.\ $(x,y,z) = (i \cdot s_v,\, j \cdot s_v,\, k \cdot s_v)$. Figure~\ref{fig:orthogonal_slices} shows three orthogonal views extracted through the volume center for visualization and quality assessment.

\subsection*{Rotation Axis Estimation}

\subsubsection*{Edge Detection and Contour Extraction}

For each axial slice $k \in [k_{\text{start}}, k_{\text{end}}]$, where the range is limited to slices in which the chamber cross-section is fully visible and a complete ellipse can be detected, we extract the 2D slice $I_k(x,y) = V_{\text{CT}}(x, y, k)$ and compute the gradient magnitude using the Scharr operator~\cite{scharr2000optimal}:
\begin{equation}
G_k(x,y) = \sqrt{(I_k * S_x)^2 + (I_k * S_y)^2}
\label{eq:gradient_magnitude}
\end{equation}
where $S_x$ and $S_y = S_x^T$ are the horizontal and vertical Scharr kernels --- transposes of each other, defined analytically to minimize angular error for $3\times3$ support --- they are not tuned per dataset. Compared to Sobel or Prewitt operators, the Scharr kernel provides better isotropic response, which yields more uniform edge detection on curved boundaries regardless of their orientation in the slice. Since the subsequent axis estimation aggregates ellipse centers across hundreds of slices and the translational alignment corrects any residual offset, the exact choice of gradient operator has limited impact on the final registration result.

\textbf{Binary edge map} via thresholding:
\begin{equation}
E_k(x,y) = \begin{cases}
1 & \text{if } G_k(x,y) > \tau \\
0 & \text{otherwise}
\end{cases}
\label{eq:binary_edge_map}
\end{equation}
The threshold $\tau$ is chosen empirically to separate chamber material from the surrounding air background; its exact value is scanner-dependent and does not affect the subsequent geometric pipeline, as only edge locations---not intensity values---are used for ellipse fitting.

\textbf{Contour extraction}: Connected components in $E_k$ are identified using standard two-pass labeling~\cite{suzuki1985topological}, and the largest $N_c = 4$ contours $\{C_i^k\}_{i=1}^4$ are retained for ellipse fitting. Four contours arise because each of the two cylindrical boundaries (inner and outer wall surface) produces two gradient peaks --- one on each side of the material transition --- yielding two detectable edge contours per boundary. To reduce the risk of retaining artefact-induced contours (e.g., beam hardening streaks), each fitted ellipse is validated by its aspect ratio; contours yielding degenerate fits are discarded. Residual outliers are handled by the subsequent RANSAC step.

\begin{figure}[ht]
\centering
\includegraphics[width=\textwidth]{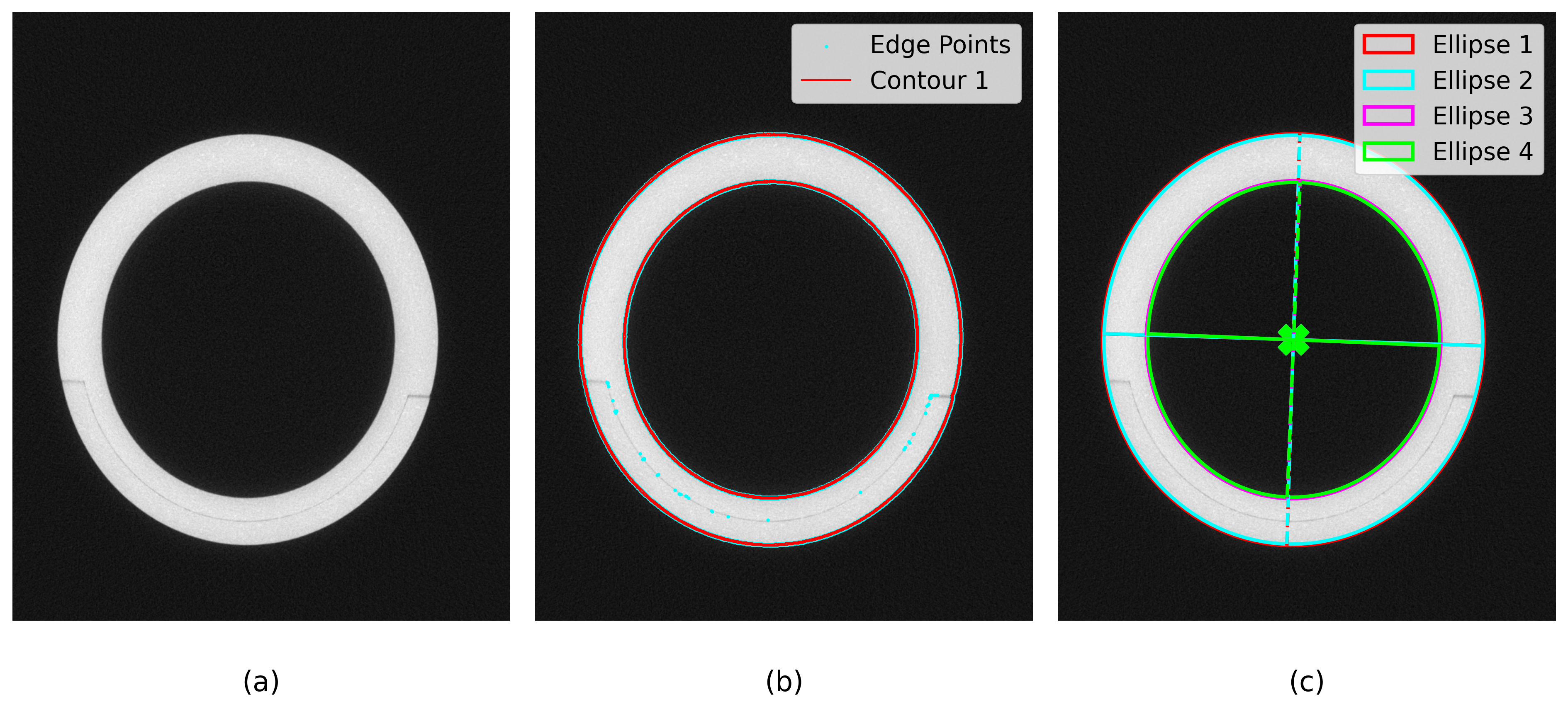}
\caption{Ellipse fitting pipeline on a single CT slice: (a) original CT slice $I_k$, (b) detected edge points (cyan) where gradient magnitude $G_k > \tau$ and extracted contours (red) from connected components, (c) fitted ellipses with major (solid) and minor (dashed) axes. Four ellipses correspond to inner and outer chamber surfaces.}
\label{fig:edge_detection}
\end{figure}

\subsubsection*{Ellipse Fitting with Sub-pixel Accuracy}

For each contour $C_i^k$ represented as a set of pixel coordinates $\{(x_j, y_j)\}_{j=1}^{N_p}$, we fit an ellipse using the direct least-squares method of Fitzgibbon et al.~\cite{fitzgibbon1999direct}. The ellipse is represented as a general conic $ax^2 + bxy + cy^2 + dx + ey + f = 0$, parameterized by $\mathbf{a} = [a, b, c, d, e, f]^T$. Since conic coefficients are defined up to scale, the ellipticity condition $b^2 - 4ac < 0$ is turned into the equality constraint $4ac - b^2 = 1$ (i.e., $b^2 - 4ac = -1$), which simultaneously fixes the scale ambiguity and guarantees a unique elliptic solution. This constraint is encoded in the matrix $\mathbf{C}$ below.

\textbf{Design matrix} $\mathbf{D} \in \mathbb{R}^{N_p \times 6}$ and \textbf{constraint matrix} $\mathbf{C}$:
\begin{equation}
\mathbf{D} = \begin{bmatrix}
x_1^2 & x_1 y_1 & y_1^2 & x_1 & y_1 & 1 \\
x_2^2 & x_2 y_2 & y_2^2 & x_2 & y_2 & 1 \\
\vdots & \vdots & \vdots & \vdots & \vdots & \vdots \\
x_{N_p}^2 & x_{N_p} y_{N_p} & y_{N_p}^2 & x_{N_p} & y_{N_p} & 1
\end{bmatrix}, \quad
\mathbf{C} = \begin{bmatrix}
0 & 0 & 2 & 0 & 0 & 0 \\
0 & -1 & 0 & 0 & 0 & 0 \\
2 & 0 & 0 & 0 & 0 & 0 \\
0 & 0 & 0 & 0 & 0 & 0 \\
0 & 0 & 0 & 0 & 0 & 0 \\
0 & 0 & 0 & 0 & 0 & 0
\end{bmatrix}
\label{eq:design_constraint_matrices}
\end{equation}

\textbf{Constrained optimization}:
\begin{equation}
\min_{\mathbf{a}} \mathbf{a}^T \mathbf{D}^T \mathbf{D} \mathbf{a} \quad \text{subject to} \quad \mathbf{a}^T \mathbf{C} \mathbf{a} = 1
\label{eq:constrained_optimization}
\end{equation}
where $\mathbf{a} = [a, b, c, d, e, f]^T$ contains the conic parameters.

\textbf{Ellipse center extraction}:
\begin{equation}
\mathbf{c}_i^k = \begin{bmatrix} c_x \\ c_y \end{bmatrix} = \frac{1}{b^2 - 4ac} \begin{bmatrix} 2cd - be \\ 2ae - bd \end{bmatrix}
\label{eq:ellipse_center}
\end{equation}
where $f$ does not appear, as it cancels algebraically when solving the gradient conditions of the conic.

\textbf{Sub-pixel accuracy}: Unlike pixel-based methods (e.g., Hough transform), least-squares ellipse fitting operates on continuous coordinates and provides \textbf{sub-pixel localization} of ellipse centers~\cite{fitzgibbon1999direct}. The fitting minimizes algebraic distance over all contour points simultaneously, achieving typical center localization accuracy of $\pm 0.1$ pixels. This sub-pixel precision is critical for accurate 3D axis estimation, as pixel-level quantization would introduce systematic errors of order $s_v$ in the reconstructed axis orientation. The geometric consistency analysis in Figure~\ref{fig:ewma_stability} provides empirical confirmation: ellipse center spacing varies with $\sigma < 0.01$ px across all slices, well below one pixel.

For a cylindrical object with radius $R$ tilted at angle $\theta$ from the imaging plane normal, the ellipse centers trace a line whose slope directly determines $\theta$. Sub-pixel center localization enables angular resolution:
\begin{equation}
\Delta \theta \approx \frac{\Delta c}{R}
\label{eq:angular_resolution}
\end{equation}
where $\Delta c \ll s_v$ is the sub-pixel localization error.

\subsubsection*{3D Point Cloud Construction}

Ellipse centers across all processed slices form a 3D point cloud:
\begin{equation}
\mathcal{P} = \{\mathbf{p}_i\}_{i=1}^{N_{\text{total}}}, \quad \mathbf{p}_i = \begin{bmatrix} c_x^i & c_y^i & z_i \end{bmatrix}^T, \quad z_i = k_i \cdot s_v
\label{eq:point_cloud}
\end{equation}
with $k_i$ being the slice index and $c_x^i, c_y^i$ the ellipse center in slice $k_i$. Typically, $N_{\text{total}} = 4 \times (k_{\text{end}} - k_{\text{start}} + 1)$ points (4 ellipses per slice).

\subsubsection*{Geometric Consistency Validation}

\begin{figure}[ht]
\centering
\includegraphics[width=0.6\textwidth]{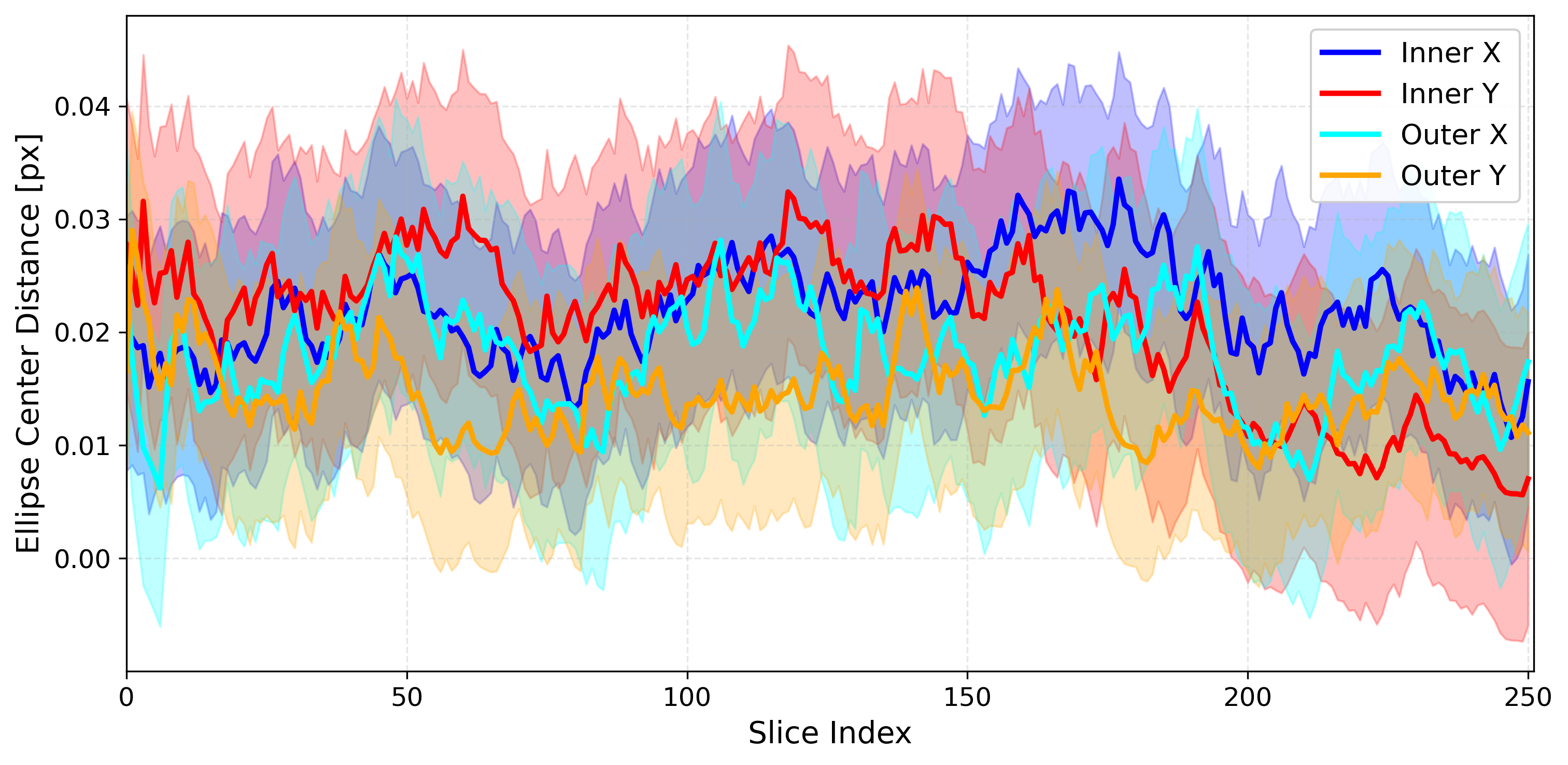}
\caption{Geometric consistency validation: distances between inner and outer ellipse centers across slices with EWMA smoothing ($\alpha=0.15$) and standard deviation bands.}
\label{fig:ewma_stability}
\end{figure}

Before proceeding to axis fitting, we validate the geometric consistency of the detected ellipses. For cylindrical objects with concentric inner and outer surfaces, the spacing between corresponding ellipse centers should remain approximately constant across slices. We compute distances between inner ellipse pairs and outer ellipse pairs:
\begin{equation}
d_{\text{inner}}^k = \|\mathbf{c}_1^k - \mathbf{c}_2^k\|, \qquad d_{\text{outer}}^k = \|\mathbf{c}_3^k - \mathbf{c}_4^k\|
\label{eq:ellipse_distances}
\end{equation}
To assess stability, we apply exponentially weighted moving average (EWMA) smoothing with a decay factor $\alpha = 0.15$, chosen heuristically to suppress high-frequency noise while preserving slow trends:
\begin{equation}
\text{EWMA}_k = \alpha \cdot d^k + (1-\alpha) \cdot \text{EWMA}_{k-1}
\label{eq:ewma}
\end{equation}
Both spacings remain stable throughout the scan ($\sigma_{\text{inner}} = 0.008$ px, $\sigma_{\text{outer}} = 0.006$ px), as shown in Figure~\ref{fig:ewma_stability}. The slight decline beyond slice 200 indicates the chamber approaches the scan boundary.

\subsubsection*{RANSAC-based Line Fitting}

\begin{figure}[ht]
\centering
\includegraphics[width=\textwidth]{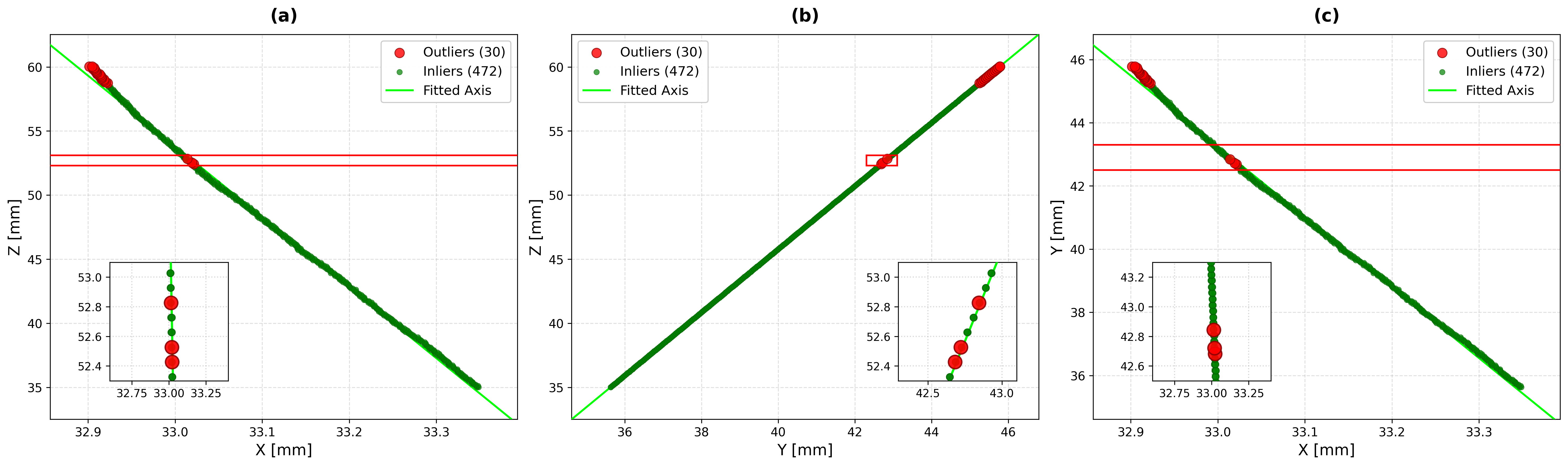}
\caption{RANSAC outlier removal visualized through 2D projections: (a) XZ projection (side view), (b) YZ projection (front view), (c) XY projection (top view). Red circles are outliers rejected by RANSAC, green circles are inliers. The line shows the PCA-fitted axis estimated from the inlier set. Inset windows (red rectangles, $0.6\times0.6$ mm) show sub-pixel precision of the recovered axis.}
\label{fig:ransac}
\end{figure}

We apply RANSAC~\cite{fischler1981random} to fit a 3D line through $\mathcal{P}$, rejecting outliers arising from incorrectly fitted ellipses or partial volume effects at slice boundaries. A point $\mathbf{p}_i$ is classified as an inlier if its perpendicular distance to the candidate line falls below threshold $\delta_{\text{RANSAC}}$:
\begin{equation}
d_\perp(\mathbf{p}_i) = \frac{\|(\mathbf{p}_i - \mathbf{p}_0) \times \mathbf{d}\|}{\|\mathbf{d}\|} < \delta_{\text{RANSAC}}
\label{eq:ransac_distance}
\end{equation}
where $\mathbf{p}_0 \in \mathbb{R}^3$ is a point on the line and $\mathbf{d} \in \mathbb{R}^3$ is its direction vector. We use $\delta_{\text{RANSAC}} = 0.1 \, s_v$ (one tenth of the voxel size). Figure~\ref{fig:ransac} shows the resulting inlier/outlier classification, typically achieving 94.0\% inliers.

\subsubsection*{PCA-based Axis Extraction}

We apply PCA to the inlier set $\mathcal{P}_{\text{inlier}} = \{\mathbf{p}_i : d_\perp(\mathbf{p}_i) < \delta_{\text{RANSAC}}\}$ to determine the principal axis direction.

\textbf{Mean position} and \textbf{covariance matrix}:
\begin{equation}
\boldsymbol{\mu} = \frac{1}{N_{\text{inlier}}} \sum_{i=1}^{N_{\text{inlier}}} \mathbf{p}_i, \quad
\boldsymbol{\Sigma} = \frac{1}{N_{\text{inlier}}} \sum_{i=1}^{N_{\text{inlier}}} (\mathbf{p}_i - \boldsymbol{\mu})(\mathbf{p}_i - \boldsymbol{\mu})^T
\label{eq:mean_covariance}
\end{equation}

\textbf{Eigenvalue decomposition} and \textbf{rotation axis direction}: solving $\boldsymbol{\Sigma} \mathbf{v}_j = \lambda_j \mathbf{v}_j$ for $j=1,2,3$ with $\lambda_1 \geq \lambda_2 \geq \lambda_3$, the principal component $\mathbf{v}_1$ (largest eigenvalue) gives the axis direction:
\begin{equation}
\mathbf{d}_{\text{axis}} = \mathbf{v}_1 = \begin{bmatrix} d_x & d_y & d_z \end{bmatrix}^T
\label{eq:axis_direction}
\end{equation}

\begin{figure}[ht]
\centering
\includegraphics[width=0.34\textwidth]{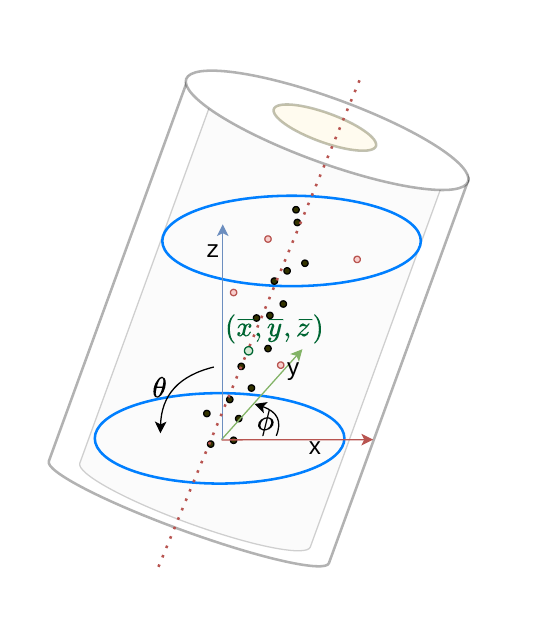}
\caption{Schematic illustration of the PCA-based axis estimation. Black dots are inlier ellipse centers, red circles are RANSAC outliers. The green point marks the centroid $\boldsymbol{\mu} = (\bar{x}, \bar{y}, \bar{z})$, the red dashed line is the recovered rotation axis $\mathbf{d}_{\text{axis}}$, and the angles $\theta$ (tilt) and $\phi$ (orientation) characterize the axis direction in the CT coordinate frame.}
\label{fig:pca_axis}
\end{figure}

\textbf{Tilt angle} $\theta$ and \textbf{orientation angle} $\phi$ (as illustrated in Figure~\ref{fig:pca_axis}):
\begin{equation}
\theta = \arccos(d_z), \quad \phi = \arctan2(d_y, d_x)
\label{eq:tilt_orientation_angles}
\end{equation}
assuming $\|\mathbf{d}_{\text{axis}}\| = 1$ after normalization. If $\phi < 0$, we add $2\pi$ to obtain $\phi \in [0, 2\pi)$.

\subsection*{Translational Alignment}

\subsubsection*{Overview}

Given the estimated rotation axis $\mathbf{d}_{\text{axis}}$ and a point on the axis $\mathbf{p}_0$, we seek the translational offset $\mathbf{t} \in \mathbb{R}^3$ that maximizes the volumetric overlap between the CAD model and the CT scan. We decompose the search into two sequential stages: alignment along the rotation axis (1D) and alignment in the perpendicular plane (2D).

The overlap score is defined as the sum of CT voxel intensities within the voxelized CAD model region:
\begin{equation}
S(\mathbf{t}) = \sum_{(i,j,k) \in M_{\text{CAD}}(\mathbf{t})} V_{\text{CT}}(i,j,k)
\label{eq:overlap_score}
\end{equation}
where $M_{\text{CAD}}(\mathbf{t})$ denotes the set of voxel indices occupied by the CAD mesh translated by $\mathbf{t}$. CT intensity values are normalized to $[0, 1]$ prior to scoring; this ensures non-negative contributions from all voxels and, importantly, prevents negative air-region intensities from distorting the shape of the score profile used for polynomial fitting.

\subsubsection*{Alignment Along the Rotation Axis}

The axis estimation yields the orientation of the rotation axis but not the position of the CAD model along it --- the axis is infinite and the chamber can sit at any axial offset. We therefore sample the score $S$ over a range of candidate positions along the axis direction, displacing the CAD model by a scalar offset $\lambda_\parallel$ [mm]:
\begin{equation}
S_{\parallel}(\lambda_\parallel) = S(\lambda_\parallel \cdot \mathbf{d}_{\text{axis}}), \quad
\hat{S}_{\parallel}(\lambda_\parallel) = a\lambda_\parallel^2 + b\lambda_\parallel + c
\label{eq:score_along_axis}
\end{equation}
The samples are fitted with a second-degree polynomial and the optimal offset is obtained analytically from $\hat{S}_{\parallel}'(\lambda_\parallel) = 0$. Sampling points are symmetrically balanced around the observed score maximum, retaining at least 6 points on each side, yielding 13 sample points; increasing this number was not found to improve the result, as the score profile is smooth and well-approximated by a parabola near the optimum. An example result is shown in Figure~\ref{fig:score_parallel_fit}.
\begin{equation}
\lambda_\parallel^* = -\frac{b}{2a}, \qquad \mathbf{t}_{\parallel} = \lambda_\parallel^* \cdot \mathbf{d}_{\text{axis}}
\label{eq:axial_optimum}
\end{equation}

\subsubsection*{Alignment Perpendicular to the Rotation Axis}

\begin{figure}[ht]
\centering
\begin{subfigure}[b]{0.48\textwidth}
    \centering
    \includegraphics[width=\textwidth]{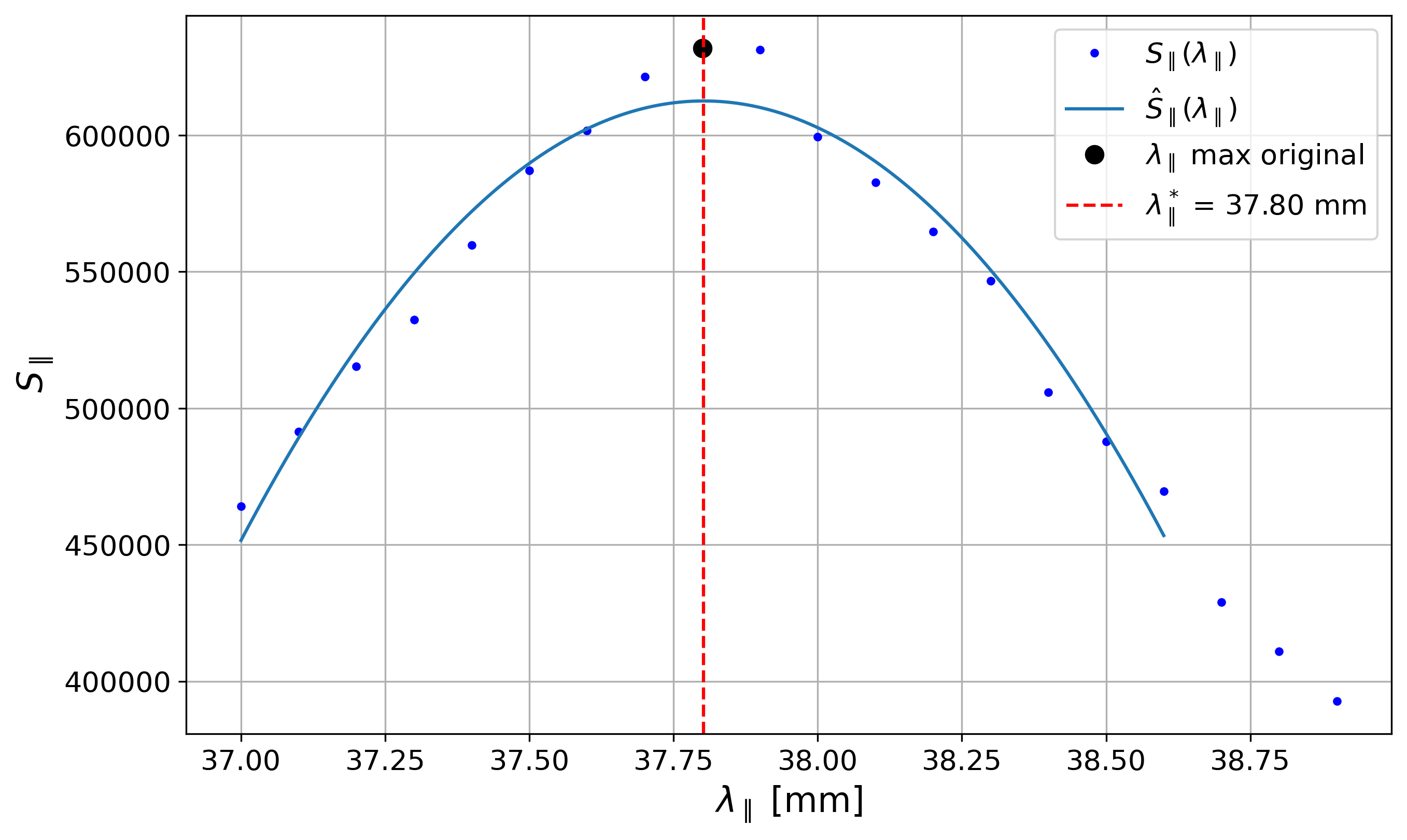}
    \caption{Score $S_\parallel(\lambda_\parallel)$ with second-degree polynomial fit. Red dashed line indicates the analytical optimum $\lambda_\parallel^*$.}
    \label{fig:score_parallel_fit}
\end{subfigure}
\hfill
\begin{subfigure}[b]{0.48\textwidth}
    \centering
    \includegraphics[width=\textwidth]{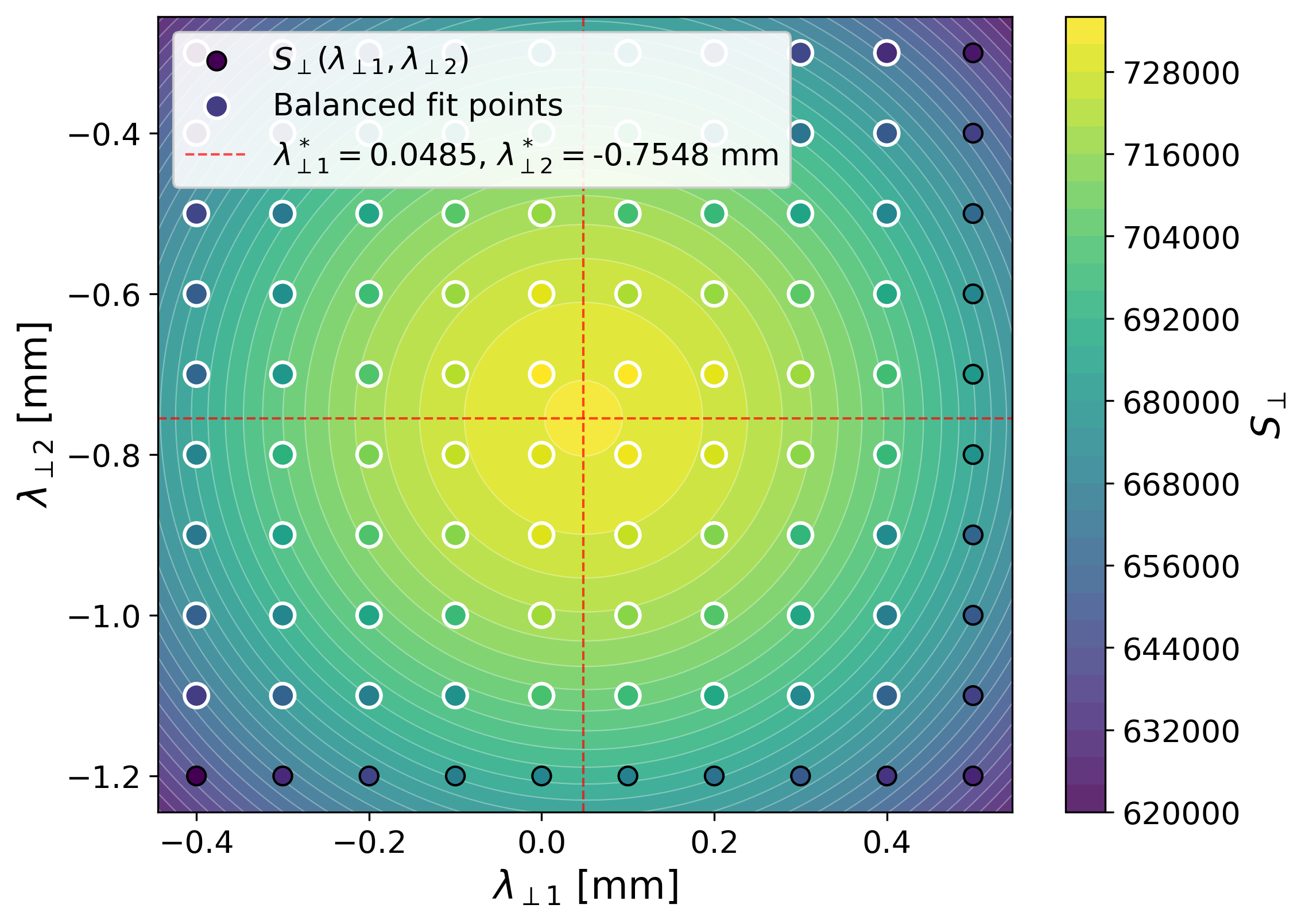}
    \caption{Score $S_\perp(\lambda_{\perp 1}, \lambda_{\perp 2})$ with fitted paraboloid $\hat{S}_\perp$. Red dashed lines mark the analytical optimum $(\lambda_{\perp 1}^*, \lambda_{\perp 2}^*)$.}
    \label{fig:score_perp_fit}
\end{subfigure}
\caption{Polynomial fitting for translational alignment.}
\label{fig:translation_fitting}
\end{figure}

Although the rotation axis direction is known, the reference point $\mathbf{p}_0$ --- the centroid of the ellipse center cloud --- may not coincide exactly with the geometric center of the CAD model. In theory they should match, but in practice asymmetric noise, partial volume effects, and beam hardening artefacts can introduce a small systematic bias in the detected edge positions, shifting the centroid from the true axis. A residual perpendicular offset therefore remains after axial alignment and must be corrected. We construct an orthonormal basis $\{\mathbf{d}_{\perp 1}, \mathbf{d}_{\perp 2}\}$ spanning the plane perpendicular to $\mathbf{d}_{\text{axis}}$ and sample the score over a 2D grid of candidate offsets $\lambda_{\perp 1}, \lambda_{\perp 2}$ [mm]:
\begin{equation}
S_{\perp}(\lambda_{\perp 1}, \lambda_{\perp 2}) = S(\mathbf{t}_{\parallel} + \lambda_{\perp 1}\,\mathbf{d}_{\perp 1} + \lambda_{\perp 2}\,\mathbf{d}_{\perp 2})
\end{equation}
\begin{equation}
\hat{S}_{\perp} = c_{00} + c_{10}\lambda_{\perp 1} + c_{01}\lambda_{\perp 2} + c_{20}\lambda_{\perp 1}^2 + c_{11}\lambda_{\perp 1}\lambda_{\perp 2} + c_{02}\lambda_{\perp 2}^2
\label{eq:score_perp}
\end{equation}
The samples are fitted with a second-degree bivariate polynomial and the optimal offset is found analytically from $\nabla \hat{S}_{\perp} = \mathbf{0}$. Sampling points are symmetrically balanced around the observed score maximum in both dimensions, retaining at least 3 points on each side, yielding a $7 \times 7$ grid of 49 sample points --- sufficient to overdetermine the 6-coefficient bivariate polynomial; as in the 1D case, increasing this number was not found to affect the result. An example result is shown in Figure~\ref{fig:score_perp_fit}.
\begin{equation}
\begin{bmatrix} \lambda_{\perp 1}^* \\ \lambda_{\perp 2}^* \end{bmatrix}
= -\begin{bmatrix} 2c_{20} & c_{11} \\ c_{11} & 2c_{02} \end{bmatrix}^{-1}
\begin{bmatrix} c_{10} \\ c_{01} \end{bmatrix}, \qquad
\mathbf{t}^* = \mathbf{t}_{\parallel} + \lambda_{\perp 1}^* \mathbf{d}_{\perp 1} + \lambda_{\perp 2}^* \mathbf{d}_{\perp 2}
\label{eq:perp_optimum}
\end{equation}

\subsection*{Evaluation on Simulated Data}

Simulated CT data provide a controlled environment where ground-truth geometry is known exactly. The CAD model used to generate the simulation serves directly as the registration reference, allowing quantitative evaluation of both rotational and translational alignment accuracy.

The simulated CT data were generated using a multi-stage pipeline. Parametrized CAD models of ionization chambers were constructed using the cadquery library~\cite{cadquery}, then converted to triangular meshes using FreeCAD~\cite{freecad} with the standard meshing algorithm. Voxelization of the resulting meshes was performed using Open3D~\cite{zhou2018open3d}, with empty regions filled using an odd-even connected component strategy to correctly handle concentric boundaries. Forward projection of the CAD-derived voxel model with material-specific attenuation coefficients from NIST tables~\cite{hubbell1995tables} was performed using the CIL framework~\cite{jorgensenCoreImagingLibrary2021}; material assignments follow the physical construction of the chambers: the wall and electrode were assigned graphite, the stem and cable aluminium, and the end cap PVC attenuation coefficients. This was followed by polychromatic beam simulation according to the method of Lifton et al.~\cite{lifton2016method}, with the X-ray spectrum generated using SpekPy~\cite{poludniowski2021spekpy} to match the acquisition parameters. Realistic measurement noise was then added. Gaussian blurring with $\sigma = 0.9$ was applied to the sinogram to model the finite detector resolution, calibrated against real CT measurements of an aluminium aerosol container with known wall thickness $d = 0.3$~mm. Detector noise was modelled as additive Gaussian with $\sigma = 0.0055$, estimated from the background region of real radiographs. The ground-truth segmentation masks used for evaluation were derived directly from the input voxel model, providing exact binary references for the wall and air cavity regions.

\subsubsection*{Rotational Alignment Accuracy}

The rotation axis estimation was evaluated by comparing the axis direction recovered from ellipse trajectory analysis against the known CAD geometry, position and orientation used to generate the simulated scans. Table~\ref{tab:rotation_metrics} summarizes the angular errors across 20 simulated scans.

\begin{table}[ht]
\centering
\caption{Rotational alignment errors on simulated CT data}
\label{tab:rotation_metrics}
\begin{tabular}{lcc}
\hline
Parameter & Mean error & Std \\
\hline
Tilt $\theta$      & $0.023^\circ$ & $0.008^\circ$ \\
Orientation $\phi$ & $0.050^\circ$ & $0.021^\circ$ \\
\hline
\end{tabular}
\end{table}

Both tilt and orientation errors remain well below $0.1^\circ$, confirming that the RANSAC-PCA pipeline reliably recovers the rotation axis from ellipse centers with sub-degree precision.

\subsubsection*{Translational Alignment Accuracy}

Translational accuracy was assessed by comparing the voxelized CAD wall mesh after registration against the ground-truth segmentation mask generated directly from the voxelized CAD model used to produce the simulation. Three volumetric overlap metrics were computed: Intersection over Union (IoU/Jaccard), Dice coefficient, and Overlap coefficient~\cite{taha2015metrics}. Results are summarized in Table~\ref{tab:translation_metrics}.

\begin{table}[ht]
\centering
\caption{Translational alignment accuracy for infill and walls on simulated CT data ($N=20$)}
\label{tab:translation_metrics}
\begin{tabular}{lccccc}
\hline
 & \multicolumn{2}{c}{\textbf{Infill}} & & \multicolumn{2}{c}{\textbf{Wall}} \\
\cline{2-3} \cline{5-6}
Metric & Mean & Std & & Mean & Std \\
\hline
IoU (Jaccard)       & $0.9982$ & $0.0009$ & & $0.9985$ & $0.0008$ \\
Dice coefficient    & $0.9991$ & $0.0006$ & & $0.9992$ & $0.0006$ \\
Overlap coefficient & $0.9993$ & $0.0004$ & & $0.9995$ & $0.0005$ \\
\hline
\end{tabular}
\end{table}

\begin{figure}[ht]
\centering
\begin{subfigure}[t]{0.30\textwidth}
    \centering
    \includegraphics[width=\textwidth]{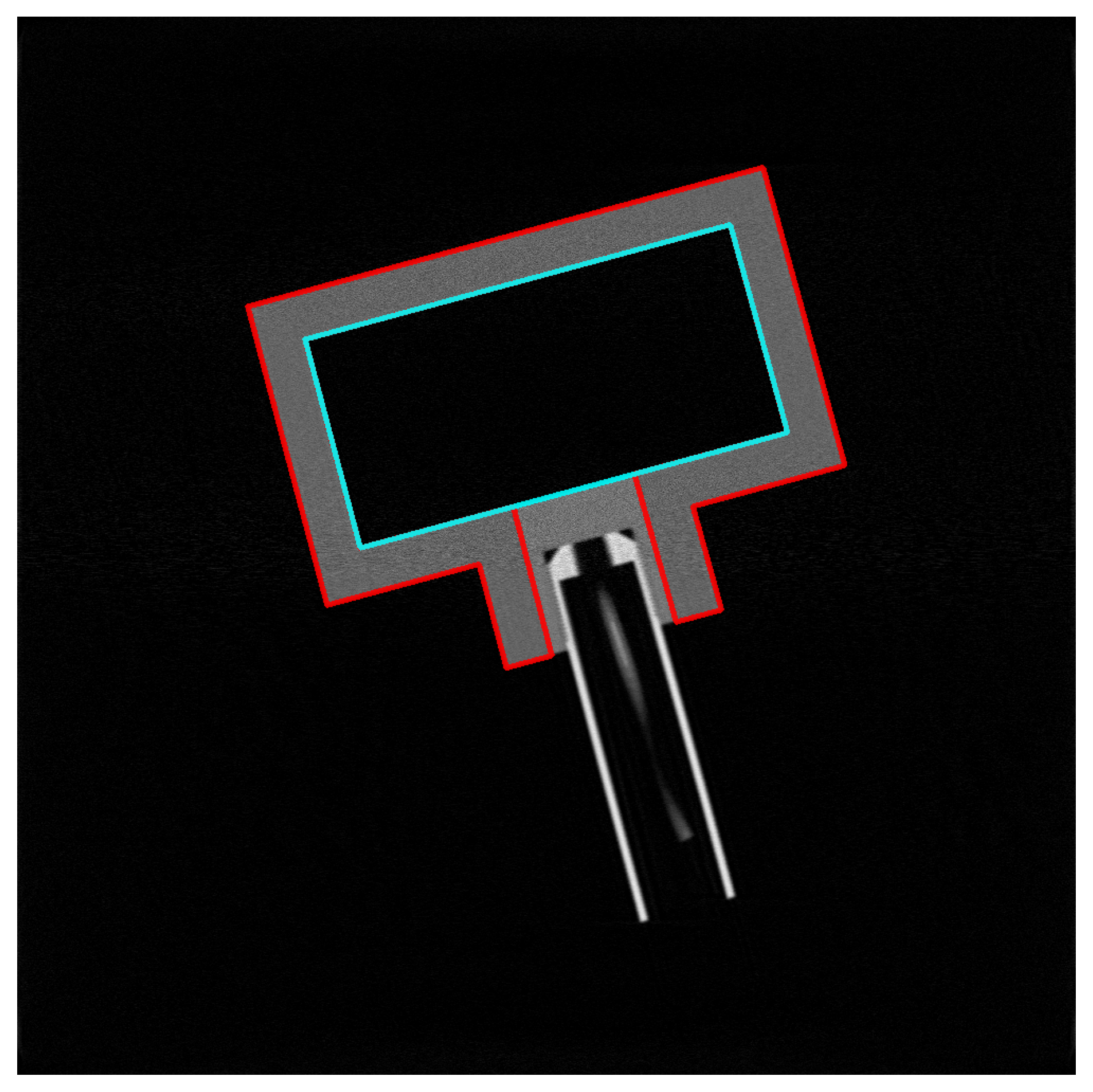}
    \caption{Registered CAD model overlaid on CT scan.}
    \label{fig:cad_ct_alignment}
\end{subfigure}
\hspace{0.02\textwidth}
\begin{subfigure}[t]{0.30\textwidth}
    \centering
    \includegraphics[width=\textwidth]{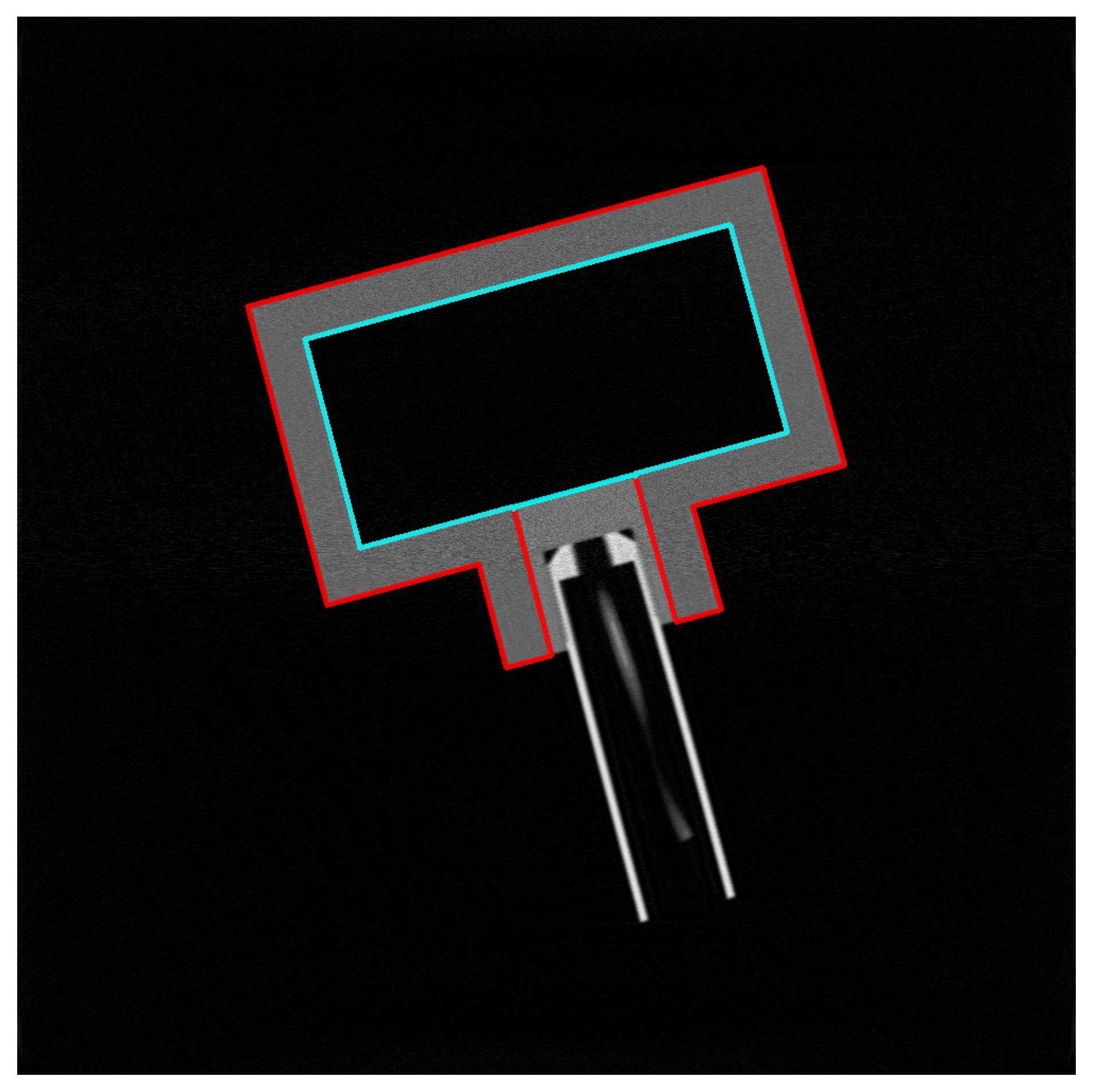}
    \caption{Ground truth CAD model used for simulation generation.}
    \label{fig:cad_mask_alignment}
\end{subfigure}
\hspace{0.02\textwidth}
\begin{subfigure}[t]{0.30\textwidth}
    \centering
    \includegraphics[width=\textwidth]{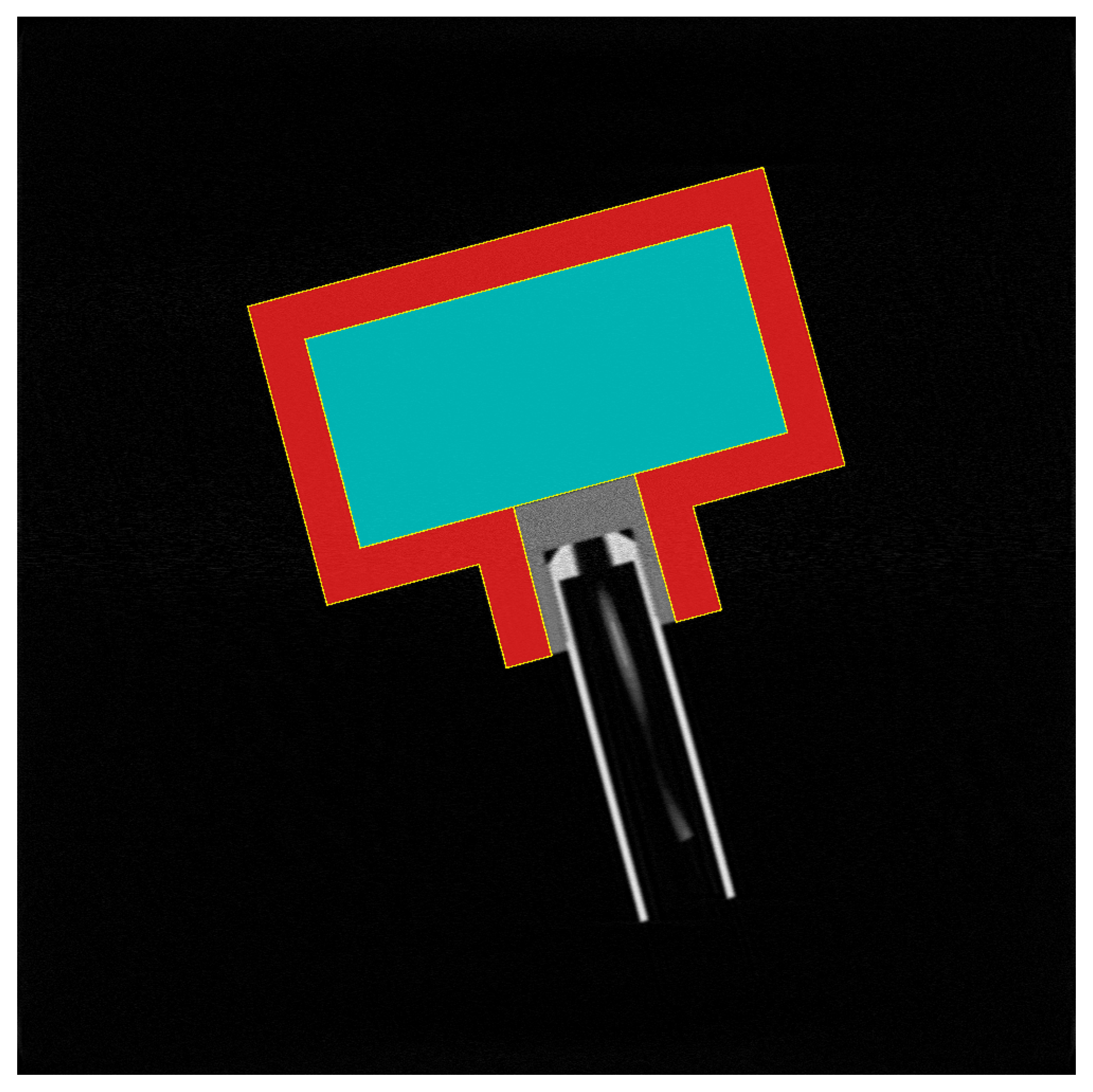}
    \caption{Segmentation masks derived from the registered CAD geometry.}
    \label{fig:cad_ct_full}
\end{subfigure}
\caption{Registration result on simulated CT data.}
\label{fig:registration_result}
\end{figure}

The near-unity overlap metrics confirm that the two-stage translational alignment achieves sub-voxel registration accuracy on simulated data. The low standard deviation across scans indicates robust and consistent performance independent of the specific scan configuration. A qualitative example of the registration result is shown in Figure~\ref{fig:registration_result}.

\subsection*{Results on Real Ionization Chambers}
\label{sec:real_chambers}

In order to verify the accuracy of the proposed registration on real CT scans, a downstream segmentation task is used. In the case of a CT scan of a real object there is no ground truth for intermediate pipeline stages: the exact rotation axis orientation, translational position, and per-voxel segmentation masks are unknown, so the registration quality cannot be assessed directly. This makes the validation approach used for simulated data not applicable. Performance of a supervised training for a downstream segmentation task can be used as a proxy for CAD registration quality. The accuracy of the registration affects the quality of extracted labels, which are decisive to the machine learning based segmentation model performance when trained in a supervised setting. In particular, using machine learning based segmentation the final chamber geometry can be predicted and compared against an external reference. Any systematic error in axis estimation, translational alignment, or segmentation will propagate into the final volumetric result.

\begin{figure}[ht]
\centering
\begin{subfigure}[b]{0.45\textwidth}
    \centering
    \includegraphics[height=8cm]{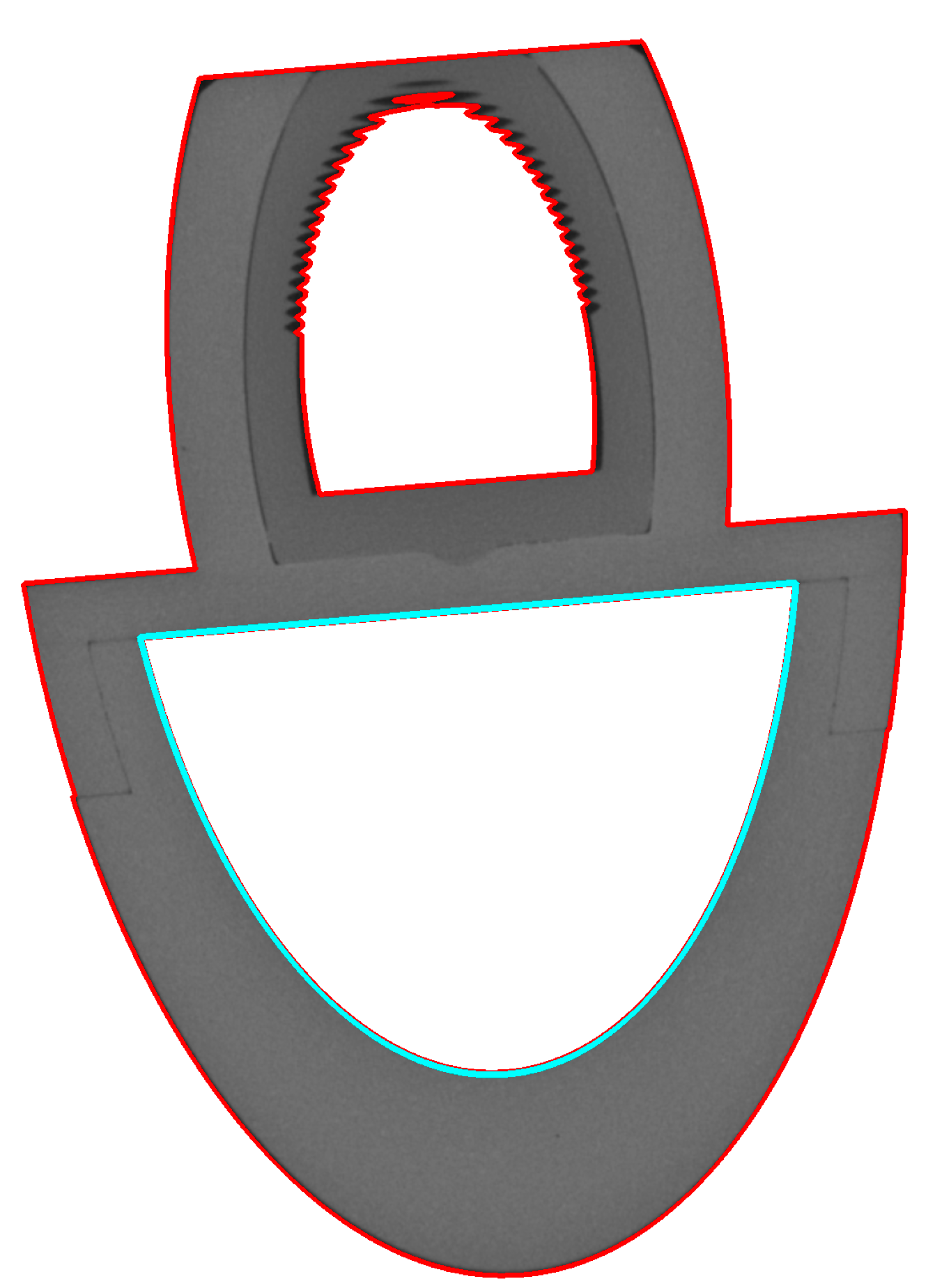}
    \caption{Example axial CT cross-section with predicted segmentation mask contours overlaid. The chamber interior fill (cyan) and wall (red) regions are shown.}
    \label{fig:real_slice_2d}
\end{subfigure}
\hfill
\begin{subfigure}[b]{0.45\textwidth}
    \centering
    \includegraphics[height=8cm]{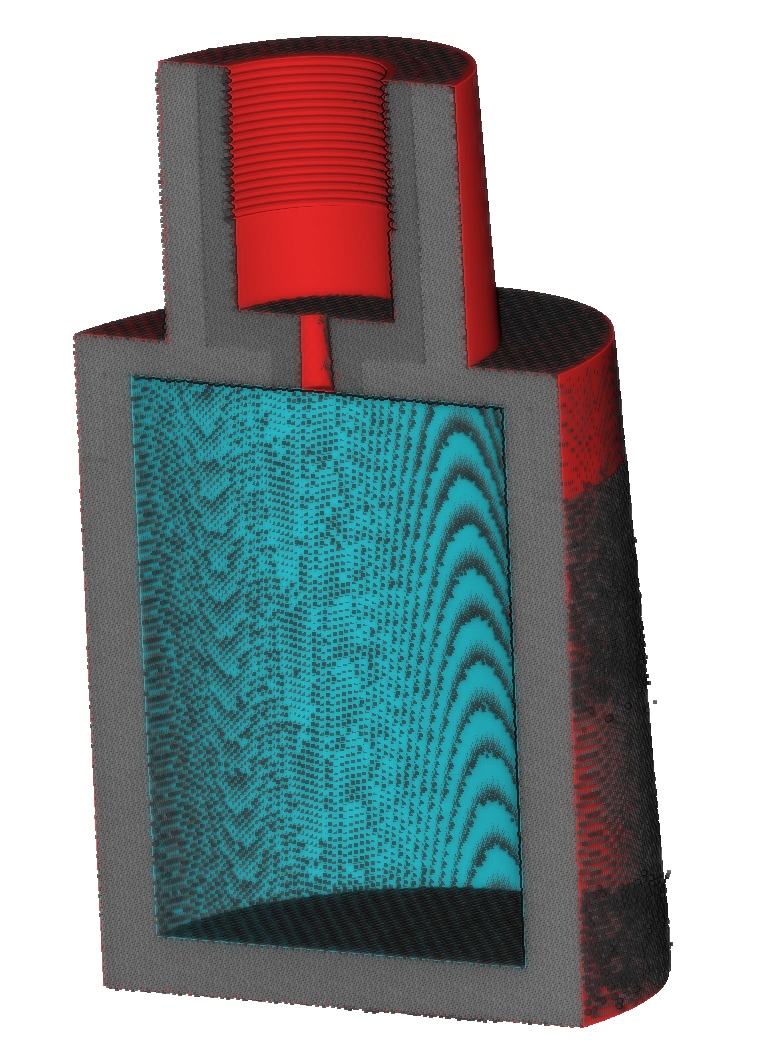}
    \caption{3D cross-sectional view of the registered CAD model overlaid on the same CT volume.}
    \label{fig:real_slice_3d}
\end{subfigure}
\caption{Registration and segmentation results for an example real ionization chamber. The CT isosurface in (b) is rendered with arbitrary intensity thresholding for visualization only.}
\label{fig:real_qualitative}
\end{figure}

The 18 ionization chambers are split into a training set ($N=9$) and a test set ($N=9$). For training chambers, the registered CAD model provides voxel-level segmentation labels used to train the supervised segmentation model. For test chambers, no CAD-derived labels are used during inference --- the CAD model serves only as a volumetric reference for end-to-end validation against CMM measurements described in the following subsection. Visual registration results are shown in Figure~\ref{fig:real_qualitative}.

\subsubsection*{Reference Measurements and CAD Model Construction}
\label{sec:reference_measurements}

All 18 chambers underwent the same two-step measurement procedure. First, optical surface scanning with an ATOS II Triple Scanner (GOM) provided a point cloud from which an initial CAD model with dimensional annotations was derived. Second, a DEA GLOBAL Advantage CMM independently measured the physical chamber dimensions and corrected the ATOS-derived CAD model accordingly --- ensuring that $M_{\text{CAD}}$ reflects true manufactured geometry rather than the optical scan alone. The CMM also provided the reference interior volume used for validation, with a mean relative measurement accuracy of $0.027\%$ from uncertainty propagation.

CT scans of the chambers were obtained using a Nikon XT H 225ST cone-beam CT scanner. The acquisition parameters are summarised in Table~\ref{tab:ct_params}. Chambers were scanned at varying tilt angles and distances from the detector to increase geometric variability in the dataset. The wide range of projection counts reflects varying acquisition settings across chambers, chosen to further increase dataset diversity.

\begin{table}[ht]
\centering
\caption{CT acquisition parameters for ionization chamber scans}
\label{tab:ct_params}
\begin{tabular}{ll}
\hline
Parameter & Value \\
\hline
Scanner                     & Nikon XT H 225ST \\
Detector                    & $2000\times2000$ px, pixel size 0.2~mm \\
Tube voltage                & 220~kV \\
Tube current                & 135~$\mu$A \\
Filter                      & Cu, 0.25~mm \\
Source-to-detector distance & 1177.7~mm \\
Source-to-object distance   & $\approx$588.9~mm \\
Voxel size                  & $\approx$0.1~mm (isotropic) \\
Projections per scan        & 944--2866 \\
Angular step                & $0.13^\circ$--$0.38^\circ$ \\
\hline
\end{tabular}
\end{table}

The two sets differ only in how $M_{\text{CAD}}$ is used: for training chambers, it is registered to the CT scan to generate segmentation labels; for test chambers, it serves solely as a volumetric reference against which the model's predictions are compared --- no registration is performed on test data.

\subsubsection*{Segmentation Model}

The registered CAD-based masks provide per-voxel class labels (chamber interior fill, wall) used directly as training targets. Segmentation is performed using UNet++~\cite{zhou2019unetplusplus} with an EfficientNet-B3~\cite{tan2019efficientnet} encoder. The model operates on 2D cross-sections: for each axial slice, $N=7$ consecutive slices are stacked as input channels to provide spatial context from neighbouring cross-sections. Both training and inference follow the same sliding-window scheme --- the mask is predicted only for the central slice, and a complete 3D segmentation is assembled slice by slice across the full CT volume. The natural tilt of the chambers within the scanner introduces variability in cross-sectional ellipse shape and orientation across slices, providing implicit data augmentation that improves generalization.

\subsubsection*{Volumetric Accuracy}

The predicted interior volume was compared against the CMM reference measurement. The relative volume error is defined as:
\begin{equation}
\varepsilon = \left(\frac{\hat{V}}{V_{\text{ref}}} - 1\right) \times 100\%
\end{equation}
where $\hat{V}$ is the predicted volume and $V_{\text{ref}}$ is the reference measurement. Quantitative results for all chambers are summarized in Table~\ref{tab:real_results}.

\begin{table}[ht]
\centering
\caption{Relative volume error $\varepsilon$ [\%] for real ionization chambers}
\label{tab:real_results}
\begin{tabular}{lccc}
\hline
Split & Mean [\%] & STD [\%] & Range [\%] \\
\hline
Train & $+0.04$ & $0.17$ & $[-0.18,\ +0.30]$ \\
Test  & $-0.07$ & $0.29$ & $[-0.45,\ +0.53]$ \\
\hline
\end{tabular}
\end{table}

The training set achieves a mean relative error of $+0.04\%$ with a standard deviation of $0.17\%$, indicating a negligible systematic bias. The test set yields a mean error of $-0.07\%$ with a standard deviation of $0.29\%$, confirming that the pipeline generalizes to unseen chambers. The wider spread in the test set ($[-0.45\%, +0.53\%]$) is expected: the test chambers were unseen during training, with potentially different geometries and CT acquisition conditions --- yet the error remains at a sub-percent level, confirming acceptable generalization.

The real-data accuracy is somewhat lower than the simulation-based results in Table~\ref{tab:translation_metrics} (the wall IoU standard deviation of $0.0008$ on simulated data reflects registration error alone, whereas the test set volumetric std of $0.29\%$ additionally reflects segmentation model inaccuracy). The sub-percent accuracy on real data nonetheless confirms that the CAD-to-CT registration provides sufficiently accurate geometric reference for volumetric measurement without any labeled real data.

\section*{Summary}

\subsection*{Results}

The EWMA analysis of ellipse center positions across CT slices (Figure~\ref{fig:ewma_stability}) provides direct empirical evidence of sub-voxel geometric stability. The spacing between inner ellipse centers varies with a standard deviation of $\sigma_{\text{inner}} = 0.008$ px, and between outer centers $\sigma_{\text{outer}} = 0.006$ px --- both well below one pixel. Since consecutive axial slices are separated by exactly one voxel ($s_v$), this stability implies that the ellipse fitting localizes the chamber axis trajectory with precision significantly finer than the voxel size. The resulting axis estimation errors of $0.023^\circ$ (tilt) and $0.050^\circ$ (orientation) on simulated data confirm that sub-pixel center localization translates directly into sub-degree rotational accuracy.

Translational alignment evaluated on simulated data yields IoU above $0.998$ and Dice above $0.999$ for both the chamber interior and wall --- these metrics reflect the combined effect of rotational and translational registration errors, confirming sub-voxel end-to-end positional accuracy.

On real ionization chambers, the volumetric error on the test set reaches a mean of $-0.07\%$ and standard deviation of $0.29\%$ against CMM reference measurements, where the total error reflects the combined contribution of registration inaccuracy and segmentation model error.

\subsection*{Conclusion}

We presented a two-stage geometry-based method for registering CAD models to CT scans of cylindrical ionization chambers, requiring no intensity calibration, manual annotation, or feature correspondence assumptions. The first stage estimates the 3D rotation axis from the linear trajectory of ellipse centers across CT cross-sections using RANSAC-PCA. The second stage maximizes volumetric overlap through polynomial-fitted 1D and 2D translational search.

The registered CAD geometry provides ground truth segmentation masks that enable downstream machine learning-based analysis and manufacturing tolerance validation. While the current pipeline targets the chamber interior fill and wall, the framework can be extended to additional CAD components. Two strategies are possible: independent registration of individual components exploiting their local geometry, or simultaneous alignment of the full CAD assembly where the spatial relationships between components defined in the CAD model are preserved as rigid constraints during optimization.

The method has several limitations that scope its current applicability. First, it requires the object to contain at least one rotationally symmetric element producing a detectable elliptical cross-section in CT --- objects with purely rectangular or irregular geometry are not supported. Second, the approach relies on a precise CAD model derived from high-accuracy dimensional metrology; significant deviations between the CAD geometry and the physical object (e.g., due to manufacturing tolerances or wear) will directly degrade registration quality. Third, the method assumes that complete elliptical cross-sections are visible across a sufficient range of axial slices, which may not hold for very short objects or scans with limited axial coverage. Finally, the current pipeline does not model or compensate for non-rigid deformations, restricting applicability to rigid, dimensionally stable components.

However, the general method can be extended to objects with partial symmetry or non-circular cross-sections, as long as the cross-section shape can be parametrised analytically. By extending the elliptical fit to other shapes the whole pipeline can be repurposed. Such extension, as well as extension to objects without a precise reference CAD model, remains an open direction for future work.

\section*{Acknowledgment}
This work was completed with resources provided by the Swierk Computing Centre at the National Centre for Nuclear Research.
The AI4GUM - INFOSTRATEG7/0001/2023 project is co-financed by the National Centre for Research and Development under the 7th call of the Strategic Programme of Scientific Research and Experimental Development `INFOSTRATEG - Advanced information, telecommunications and mechatronic technologies' (PL: INFOSTRATEG Zaawansowane technologie informacyjne, telekomunikacyjne i mechatroniczne), agreement No. INFOSTRATEG7/0001/2023.

\bibliographystyle{plain}
\bibliography{references_old}

\end{document}